\pdfoutput=1

\documentclass[11pt]{article}

\usepackage[]{EMNLP2022}

\usepackage{times}
\usepackage{latexsym}

\usepackage[T1]{fontenc}

\usepackage[utf8]{inputenc}

\usepackage{microtype}

\usepackage{inconsolata}

\usepackage{amsmath}
\usepackage{siunitx}
\usepackage[inline]{enumitem}
\usepackage{booktabs}
\usepackage{graphicx}
\usepackage{multirow}

\newcommand{\specials}[1]{{<}{#1}{>}}
\DeclareMathOperator*{\argmax}{argmax}

\title{Semantic Role Labeling Meets Definition Modeling:\\Using Natural Language to Describe Predicate-Argument Structures}

\author{Simone Conia$^{1}$\thanks{\ \ \ Equal contribution.} \qquad Edoardo Barba$^{1}$\footnotemark[1] \qquad Alessandro Scir\`e$^{1,2}$ \qquad Roberto Navigli$^{1}$ \AND
        Sapienza NLP Group\\
        Sapienza University of Rome\\
        $^1$\texttt{\{first.lastname\}@uniroma1.it} \And Babelscape, Italy \\ $^2$\texttt{scire@babelscape.com}}

\begin{document}
\maketitle
\begin{abstract}
One of the common traits of past and present approaches for Semantic Role Labeling (SRL) is that they rely upon discrete labels drawn from a predefined linguistic inventory to classify predicate senses and their arguments.
However, we argue this need not be the case.
In this paper, we present an approach that leverages Definition Modeling to introduce a generalized formulation of SRL as the task of describing predicate-argument structures using natural language definitions instead of discrete labels.
Our novel formulation takes a first step towards placing interpretability and flexibility foremost, and yet our experiments and analyses on PropBank-style and FrameNet-style, dependency-based and span-based SRL also demonstrate that a flexible model with an interpretable output does not necessarily come at the expense of performance.
We release our software for research purposes at \url{https://github.com/SapienzaNLP/dsrl}.
\end{abstract}

\section{Introduction}
\label{sec:introduction}
\begin{figure}[t]
    \centering
    \includegraphics[width=1.0\linewidth,trim={7.5cm 2.5cm 8cm 1.8cm},clip]{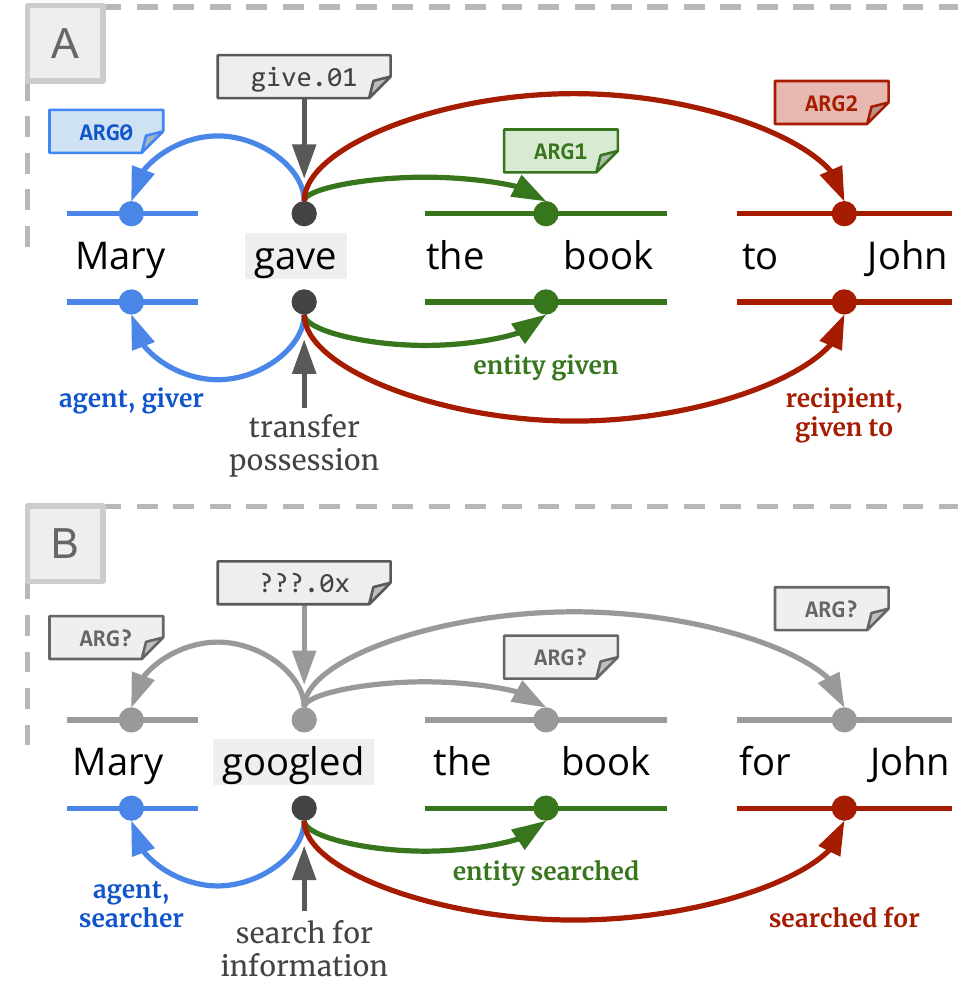}
    \caption{\textbf{A:} SRL annotations using predicate sense and semantic role labels (top) compared with their natural language definitions (bottom). \textbf{B:} the semantics of sense and role labels is undefined for out-of-inventory predicates (e.g., the inventories used for CoNLL-2009 and CoNLL-2012 do not include an entry for ``google''), but we can still use valid natural language definitions.}
    \label{fig:overview}
\end{figure}

Commonly regarded as one of the key ingredients for Natural Language Understanding~\citep{navigli-2018-ijcai}, 
Semantic Role Labeling~\cite[SRL]{gildea-jurafsky-2002-automatic} aims at identifying ``\textit{Who did What to Whom, Where, When, and How?}'' within a given sentence~\citep{marquez-etal-2008-srl}.
More precisely, for each predicate 
in the sentence, the task requires: i) selecting its most appropriate sense from a predetermined linguistic inventory; ii) identifying its arguments, i.e., those parts of the sentence that are semantically related to the predicate; and, iii) assigning a semantic role to each predicate-argument pair, as shown in Figure~\ref{fig:overview}.
Due to the potential uses of these semantically rich structures, the research community has seen steady progress in the task, and SRL has been shown to be beneficial for an increasingly wide range of applications in Natural Language Processing (NLP), such as Question Answering~\citep{shen-lapata-2007-srl-qa}, Information Extraction~\citep{christensen-etal-2011-srl-information-extraction}, Machine Translation~\citep{marcheggiani-etal-2018-srl-mt}, and Summarization~\citep{mohamed-oussalah-2021-srl-summarization}, as well as in Computer Vision for Situation Recognition~\citep{yatskar-etal-2016-srl-situation-recognition} and Video Understanding~\citep{sadhu-etal-2021-srl-video-understanding}, \textit{inter alia}.

An important yet often overlooked aspect of SRL is that, since its conception, 
the formulation of the task has generally relied upon predetermined linguistic resources, such as FrameNet~\citep{baker-etal-1998-berkeley-framenet}, PropBank~\citep{palmer-etal-2005-propbank}, VerbNet~\citep{kipper-schuler-2005-verbnet} and, more recently, VerbAtlas~\citep{di-fabio-etal-2019-verbatlas}, which provide the labels to be used for tagging predicates and their arguments with senses and semantic roles, respectively.
Therefore, to this day, 
SRL has been framed predominantly as a classification task in which systems assign \textit{discrete labels} to portions of a sentence (Figure~\ref{fig:overview}A, top).
Although recent systems have achieved impressive results on standard benchmarks~\citep{hajic-etal-2009-conll,pradhan-etal-2012-conll} in English~\citep{shi-lin-2019-simple-srl,marcheggiani-titov-2020-graph} as well as in multilingual SRL~\citep{he-etal-2019-syntax-aware-srl,conia-etal-2021-unifying-srl}, we observe and emphasize that relying upon discrete labels raises the following critical questions:
\begin{itemize}
    \item The assumption that both predicate senses and semantic roles can be unequivocally categorized into distinct classes has long been -- and still is -- at the center of numerous discussions because the boundaries between meanings are not always clear-cut \citep{tuggy-1993-ambiguity,hanks-2000-meaning}; unsurprisingly, disambiguation approaches that are not tied to specific inventories have been gaining momentum \citep{bevilacqua-etal-2020-generationary,barba-etal-2021-esc,barba-etal-2021-consec}.
    
    \item FrameNet, PropBank, and VerbNet are heterogeneous, non-overlapping resources that have led, consequently, to specialized techniques that are more effective on PropBank's rather than FrameNet's labels, or vice versa.
    
    \item Relying on any predetermined inventory hinders the ability to generalize to out-of-inventory instances. For example, some rare senses or neologisms may not be covered by the inventory of choice, which, therefore, does not define either their possible senses, or their corresponding semantic roles (Figure~\ref{fig:overview}B, top).\footnote{In several linguistic inventories, semantic role labels are defined according to specific predicate senses, i.e., they are sense-specific. This is the case for PropBank, in which core arguments (\textsc{Arg0} through \textsc{Arg5}) acquire meaning only with respect to a predicate sense, and for FrameNet, in which some frame elements are specific to the frame they belong to (e.g., \textit{Ingestible} is only defined for the frame \textit{Ingestion}). We note that this is not the case for some inventories such as VerbNet and VerbAtlas, whose semantic roles generalize across frames.}
\end{itemize}
Furthermore, recent progress in NLP at large has primarily pursued state-of-the-art results without giving much importance as to why a system may have a predilection for one particular option over the alternatives, thus making it difficult for a human to interpret their output. And SRL is no exception to this.

In this paper, instead, we put forward a generalized formulation of Definition Modeling -- the task of defining the meaning of a word or multiword expression in context -- to reframe SRL as the task of describing sentence-level semantic relations between a predicate and its arguments using natural language definitions only.
More specifically, our contributions can be summarized as follows:
\begin{enumerate}
    \item We move away from discrete labels and introduce a novel formulation of SRL that reframes the problem as the task of using natural language to describe predicate-argument structures (Figure~\ref{fig:overview}A, bottom).
    \item We propose DSRL (Descriptive Semantic Role Labeling), a simple yet effective conditional generation model to produce such natural language descriptions, dropping discrete labels while
    also demonstrating how to use these descriptions to retrieve standard SRL labels and achieve competitive or even state-of-the-art results on gold benchmarks.
    \item In contrast to previous work, our approach provides an interpretable output in natural language, can seamlessly produce descriptions according to different linguistic theories and annotation formalisms, and naturally admits descriptions for out-of-inventory instances (Figure~\ref{fig:overview}B, bottom).
    \item We provide an in-depth analysis of the strengths and pitfalls of our approach, showing where there is still room for improvement.
\end{enumerate}
We hope that our semantically-driven descriptions in natural language, free of resource-specific labels that require expert knowledge of SRL, will not only enable easier integration of sentence-level semantics into downstream applications but also provide valuable insights to NLP researchers.

\section{Related Work}
\label{sec:related-work}

\paragraph{Linguistic resources for SRL.}
As mentioned above, SRL is generally associated with a linguistic theory and a corresponding linguistic resource, which defines an inventory of predicate senses and semantic roles\footnote{More precisely, FrameNet delineates \textit{frames} and \textit{frame elements}, while VerbNet uses \textit{classes} and \textit{thematic roles}. Hereafter, for simplicity, we follow PropBank and call them \textit{senses} and \textit{semantic roles}, respectively, independently of the resource.}
\citep{baker-etal-1998-berkeley-framenet,palmer-etal-2005-propbank,kipper-schuler-2005-verbnet}.
These inventories are a rich and diverse source of expert-curated knowledge; however, aligning sense and semantic role labels across such resources using manual or automatic techniques \citep{giuglea-moschitti-2006-srl,palmer-2009-semlink,lopez-de-lacalle-etal-2014-predicate,stowe-etal-2021-semlink,conia-etal-2021-unifying-srl} is far from trivial due to their heterogeneous nature, variable degree of coverage, and different granularity.
Perhaps it is this complexity that has led researchers towards the development of approaches that are effective mainly in just one of the task ``styles'', usually PropBank-style SRL \citep[\textit{inter alia}]{marcheggiani-etal-2017-simple,cai-etal-2018-full,strubell-etal-2018-linguistically,shi-lin-2019-simple-srl,blloshmi-etal-2021-gsrl,conia-navigli-2022-probing} or FrameNet-style SRL \citep[\textit{inter alia}]{swayamdipta-etal-2017-open-sesame,peng-etal-2018-learning,lin-etal-2021-graph-semantic-parsing,pancholy-etal-2021-sister}.
To sidestep this situation, recent studies have analyzed the feasibility of moving away from rigorous linguistic resources and have looked into capturing predicate-argument relations as question-answer pairs, with promising results in the production of questions through slot-filling templates and generative models \citep{he-etal-2015-qa-srl,fitzgerald-etal-2018-qa-srl,pyatkin-etal-2021-qa-srl}.
In this paper, instead, we reframe SRL as a generalization of Definition Modeling and directly generate human-readable descriptions of the semantic relations between a predicate and its arguments, replacing discrete labels with natural language definitions to overcome the heterogeneities of linguistic inventories.

\paragraph{Recent approaches in SRL.}
Independently of the linguistic inventory of choice, given the complexity of the task, early work often employed separate systems for each step of the SRL pipeline \citep{roth-lapata-2016-neural-srl,marcheggiani-etal-2017-simple}.
However, in recent years, researchers have successfully managed to develop end-to-end approaches \cite{cai-etal-2018-full,he-etal-2018-jointly}, especially due to the increasing expressiveness of recent neural architectures.
Since then, the attention of the community has mainly focused on when syntactic features are useful \citep{strubell-etal-2018-linguistically} or can be dispensed with \citep{conia-navigli-2020-multilingual-srl}.
Further to this, several studies have also investigated the effectiveness of their proposed approaches on different annotation formalisms, namely, dependency- and span-based SRL \citep{li-etal-2019-dependency-or-span-srl,marcheggiani-titov-2020-graph}.
Most recently, sequence-to-sequence models have found renewed traction by learning to directly generate predicate-argument structures as linearized sequences \citep{blloshmi-etal-2021-gsrl,paolini-etal-2021-seq2seq-srl}.
Although the focus of our approach is to generate natural language descriptions, we stress that it can be flexibly employed to perform SRL in its traditional formulation, jointly tackling predicate sense disambiguation, argument identification and labeling in a syntax-agnostic fashion for both span- and dependency-based formalisms, the key difference being that our method also produces human-readable and, therefore, interpretable 
descriptions of the semantics of a sentence.

\paragraph{Definition Modeling.}
The task of Definition Modeling was originally concerned with producing a natural language definition for a given word and its corresponding embedding \citep{noraset-etal-2017-definition-modeling}.
The formulation of the task was later generalized to take polysemy into account, as the same word may convey different meanings depending on the context it appears in.
Although introduced a few years ago now, Definition Modeling has attracted significant interest \citep{ni-wang-2017-learning,ishiwatari-etal-2019-learning} and has found success in semantic tasks \citep{huang-etal-2019-glossbert,bevilacqua-etal-2020-generationary} such as Word Sense Disambiguation \citep[WSD]{bevilacqua-etal-2021-wsdsurvey} and Word-in-Context \citep[WiC]{pilehvar-camacho-collados-2019-wic}.
Motivated by the success of Definition Modeling, we propose a novel generalization of its formulation, in which the objective is to use natural language not only to define a target word in context but also to describe its semantically-relevant sentential constituents.

\section{Describing Predicate-Argument Structures using Natural Language}
\label{sec:methodology}
In this Section, we introduce our novel reformulation of the SRL task (Section \ref{subsec:task-formulation}), describe DSRL, a simple yet effective autoregressive approach for it (Section \ref{subsec:description-generation}), and show how to use DSRL to perform standard SRL (Section~\ref{subsec:back-to-srl}).

\subsection{Task Formulation}
\label{subsec:task-formulation}
Taking inspiration from Definition Modeling, we propose addressing predicate sense disambiguation, argument identification, and argument classification in an end-to-end fashion as the task of describing the argument structure of a predicate $p$ in a sentence $s$ by generating a natural language description $t^p$ that defines not only $p$ but also the semantic relations that connect $p$ to its arguments $a_1, a_2, \dots, a_{|A|}$, where $A$ is the set of arguments of $p$.
For example, if we consider the predicate $p =$ ``gave'' in the sentence $s =$ ``Mary \underline{gave} the book to John'', then a valid natural language description of $p$ and its argument structure could be represented as $t^p = $ ``give: \textit{transfer}. [Mary]\{\textit{giver}\} \underline{gave} [the book]\{\textit{thing given}\} [to John]\{\textit{entity given to}\}''.
Indeed, such a sequence contains i) the predicate definition for predicate sense disambiguation, ii) all the arguments of \textit{p} in \textit{s} within square brackets for argument identification, along with iii) a definition of the semantic role of each argument within curly brackets.

\subsection{Description Generation}
\label{subsec:description-generation}
To tackle our SRL formulation, we introduce a simple end-to-end autoregressive approach that, given an input sentence $s$ and a predicate $p$ in $s$, generates the natural language description $t^p$ of its argument structure.
In particular, we devise a sequence-to-sequence model whose input sequence $s^p$ is defined as follows:
\begin{align*}
s^p =\ & w_1\ \dots\ w_i\ \dots\\
      & \specials{p}\ p_1\ \dots\ p_k\ \specials{/p}\ \dots\ w_n 
\end{align*}
where $w_i$ is the $i$-th word in the original sentence $s$, while $\specials{p}$ and $\specials{/p}$ are two special markers that indicate the beginning and the end, respectively, of the predicate $p$, with $k > 1$ if $p$ is a multiword expression.
Correspondingly, we instruct the model to generate a semantically-augmented sentence $t^p$ in which: i) the sense definition of $p$ is prepended to the original sentence, ii) the arguments of $p$ are enclosed within square brackets, and, iii) each argument is followed by its semantic role definition within curly brackets.
More formally:
\begin{align*}
    t^p =\  
    & p_1\ \dots\ p_k: d_1^{p}\ \dots\ d_{k'}^{p}\ .\\ 
    & w_1 \dots \ [ w^{a_1}_{1} \dots w^{a_1}_{m_1} ]\{d_{1}^{a_1} \dots d_{m'_1}^{a_1} \} \\
    & \dots \ [ w^{a_2}_{1} \dots w^{a_2}_{m_2} ]\{ d_{1}^{a_2} \dots d_{m'_2}^{a_2} \} \\
    & ~~~ \vdots \\
    & \dots \ [ w^{a_j}_{1} \dots w^{a_j}_{m_j} ]\{d_{1}^{a_j} \dots d_{m'_j}^{a_j} \}\ \dots\ w_n
\end{align*}
where $p_i$ is the $i$-th word of the predicate $p$, $d_i^p$ is the $i$-th word of the definition of $p$, $w^{a_j}_{i}$ is the $i$-th word for the $j$-th argument of $p$, and $d_i^{a_j}$ is the $i$-th word of the definition of the semantic role for the $j$-th argument of $p$, while $k'$, $m_j$ and $m'_j$ are the length of the definition of $p$, the length of the argument $a_j$, and the length of the definition of the semantic role for $a_j$, respectively.
With this encoding, we then train our sequence-to-sequence model to learn the factorized probability $p(t^p\ |\ s^p)$ defined as follows:
\begin{align*}
    p(t^p\ |\ s^p) = p(t^p_1\ |\ s^p) \prod_{j=2}^{|t^p|} p(t^p_j\ |\ t^p_{1: j - 1}, s^p)
\end{align*}
by minimizing the cross-entropy loss with respect to the generated natural language description.

\subsection{From SRL to Natural Language and Back}
\label{subsec:back-to-srl}
Given a dataset annotated with predicate sense and role labels from an inventory that defines such labels in natural language, we note that it is always possible to convert such a dataset to our formulation.\footnote{We also note that dependency-based annotations can be seen as span-based annotations and, thus, used directly as arguments in our natural language descriptions.}
Moreover, although the main objective of our approach is to generate an output sequence that describes sentence-level semantics, in several scenarios, it is still useful to work with discrete labels for predicate senses and semantic roles, e.g., to assess the quality of the generated structures on gold benchmarks with their standard metrics.
We stress that our formulation generalizes standard SRL; casting the descriptions generated by our model to standard SRL labels is only possible if the label inventory of choice defines a suitable sense for the target predicate, which is not the case in Figure~\ref{fig:overview}B (top) as the verb ``to google'' is not covered by PropBank.
If the predicate is covered by the inventory, we can easily select the sense or the role label $\bar{y}$ whose natural language description $d^{\bar{y}}$ is most similar to the definition $d^\cdot$ generated for the predicate $p$ or for one of its arguments $a_j$.
We select $\bar{y}$ as follows:
\begin{equation*}
    \bar{y} = \argmax_{y \in Y} \sigma(f(d^y), f(d^\cdot))
\end{equation*}
where $\sigma(\cdot)$ is a similarity function (e.g., cosine similarity), $f(\cdot)$ provides a vector representation of a definition, $Y$ is the set of labels, and $d^y$ is the definition of $y$ as provided by the inventory of choice.
We note that, for simplicity, we do not apply any post-processing to enforce the validity of the generated output, leaving more complex strategies (e.g., constrained decoding) as future work.

\section{Experiments and Results}

\subsection{Data}
We train and evaluate DSRL on three widely adopted benchmarks for English SRL, namely: i) CoNLL-2009 \citep{hajic-etal-2009-conll} for dependency-based PropBank-style SRL, ii) CoNLL-2012 \citep{pradhan-etal-2012-conll} for span-based PropBank-style SRL, and iii) FrameNet 1.7 \citep{baker-etal-1998-berkeley-framenet} for span-based FrameNet-style SRL. 
While CoNLL-2009 is a collection of finance-related news from the Wall Street Journal, CoNLL-2012 is a more heterogeneous corpus comprising news, conversations, and magazine articles.
FrameNet 1.7, instead, provides a relatively small dataset of annotated documents; following the literature \citep{swayamdipta-etal-2017-open-sesame,peng-etal-2018-learning}, we include in the training set ``exemplar'' sentences extracted from partially annotated usage examples from the lexicon itself.
We provide a broader look at the characteristics of each dataset in Appendix~\ref{app:data-stats} and further details about semantic role definitions in Appendix~\ref{app:argument-modifiers}.

\subsection{Implementation Details}
We implement DSRL using Sunglasses.ai's Classy.\footnote{\url{https://github.com/sunglasses-ai/classy}}
As our underlying sequence-to-sequence model, we use BART-large~\cite{lewis-etal-2020-bart}, a Transformer-based neural network (400M parameters) pretrained with denoising objectives on massive amounts of unlabeled text.\footnote{We use the model's weights available from Huggingface Transformers~\citet{wolf-etal-2020-transformers}.} 
We do not modify its architecture except for the embedding layer, where we add the special tokens used to indicate predicates and their arguments,\footnote{See Appendix~\ref{app:special-tokens} for further details on the special tokens.} as described in Section \ref{subsec:description-generation}. 
We train our model using RAdam~\cite{liu2019variance} as the optimizer for a maximum of \num{500000} steps with a batch size of \num{2048} tokens and a standard learning rate of $10^{-5}$. 
We measure the F1 score on the validation set at the end of each training epoch, adopting an early stopping strategy to interrupt the training process if the F1 score does not improve for \num{10} consecutive epochs.
We do not modify any of the hyperparameters of BART compared to its pretraining phase, and, more generally, we do not run any hyperparameter search due to the cost of fine-tuning the language model.
The training process is carried out on a single GPU (a GeForce RTX 3090) and requires about 10 hours for FrameNet, 15 for CoNLL-2009 and 20 for CoNLL-2012.

We recall that, in order to evaluate our system with standard scoring scripts,\footnote{\href{https://ufal.mff.cuni.cz/conll2009-st/eval09.pl}{eval09.pl} for CoNLL-2009, \href{http://www.lsi.upc.es/\%7Esrlconll/srl-eval.pl}{srl-eval.pl} for CoNLL-2012, and \href{https://github.com/Noahs-ARK/semafor/blob/master/scripts/scoring/fnSemScore_modified.pl}{fnSemScore.pl} for FrameNet.} we have to cast our descriptions to the discrete labels of the target inventory (see Section \ref{subsec:back-to-srl}).
For this step, we compute the cosine similarity between the representation of a generated description and those of the possible senses or roles, using the sentence-level embeddings of SimCSE~\cite{gao-etal-2021-simcse}.\footnote{\texttt{princeton-nlp/sup-simcse-roberta-base}.}

\subsection{Comparison Systems}
We compare our results with the current state of the art in PropBank-style and FrameNet-style SRL. Following standard practice in PropBank-based SRL, we report the results achieved by our system using gold pre-identified (but not disambiguated) predicates, i.e., the position of a predicate (but not its sense label) is given as input to the system.

\paragraph{PropBank-style SRL.}
We consider \citet{li-etal-2019-dependency-or-span-srl}, who first quantified the benefits of contextualized word representations in both dependency- and span-based PropBank-style SRL, later surpassed by \citet{shi-lin-2019-simple-srl}, who used BERT instead of ELMo, and \citet{conia-navigli-2020-multilingual-srl}, who designed and took advantage of complex language-agnostic components.
We also take into account some studies for PropBank-style SRL that found success by leveraging syntactic features such as \citet{he-etal-2019-syntax-aware-srl}, who devised a strategy to cleverly prune a sentence based on its syntactic dependency tree, and \citet{marcheggiani-titov-2020-graph}, who exploited graph convolutional networks to encode syntactic relations.
Most recently, \citet{blloshmi-etal-2021-gsrl} proposed a simple and general approach to tackle SRL as a sequence-to-sequence task, in which, however, a system is still required to generate a linearized sequence of discrete labels.

\paragraph{FrameNet-style SRL.}
Although the research community has generally focused on PropBank-style SRL, especially due to the widespread adoption of PropBank in several CoNLL tasks \citep{carreras-marquez-2005-conll,surdeanu-etal-2008-conll,hajic-etal-2009-conll,pradhan-etal-2012-conll} and in other resources such as Abstract Meaning Representation \citep[AMR]{banarescu-etal-2013-amr}, FrameNet-style SRL has also been at the center of notable studies such as \citet{swayamdipta-etal-2017-open-sesame}, who investigated the effect of joint learning of syntactic and semantic features, and \citet{peng-etal-2018-learning}, who instead showed the advantages of learning from disjoint data sources.
Finally, we also consider recent work by \citet{pancholy-etal-2021-sister}, who developed a data augmentation strategy using frame relations, and the above-mentioned \citet{marcheggiani-titov-2020-graph}, who introduced a graph-based neural architecture to tackle FrameNet-style SRL.

\subsection{Main Results}
Here, we first evaluate the robustness of DSRL in achieving strong or even state-of-the-art results on standard benchmarks, and then its flexibility in performing dependency- and span-based, PropBank- and FrameNet-style SRL.
Remarkably, our model achieves even better results when jointly trained on dissimilar annotation formalisms and linguistic resources, despite their heterogeneous characteristics.

\begin{table}[t]
    \centering
    \resizebox{1.0\linewidth}{!}{%
    \begin{tabular}{lcccc}
        \toprule
        
        
        
         \textbf{CoNLL-2009} & & \textbf{\textit{~P~}} & \textbf{\textit{~R~}} & \textbf{F1} \\
        
        \midrule
        
        \citet{li-etal-2019-dependency-or-span-srl} &&
        89.6 & 91.2 & 90.4 \\
        
        \citet{he-etal-2019-syntax-aware-srl} &&
        90.4 & 91.3 & 90.9 \\
        
        \citet{shi-lin-2019-simple-srl} &&
        92.4 & 92.3 & 92.4 \\
        
        \citet{conia-navigli-2020-multilingual-srl} &&
        92.5 & 92.7 & 92.6 \\
        
        \citet{fei-etal-2021-srl-transition-based} && -- & -- & 92.2 \\
        
        \citet{blloshmi-etal-2021-gsrl} &&
        92.9 & 92.0 & 92.4 \\

        \citet{zhang-etal-2022-label-definitions} &&
        93.0 & 91.0 & 92.0 \\
        
        This work$_\textrm{ CoNLL-2009}$ &&
        92.9 & 92.1 & 92.5 \\
        
        
        This work$_\textrm{ ALL}$ &&
        92.3 & 92.4 & 92.4 \\
        
        \bottomrule
        
    \end{tabular}
    }
    \caption{Results (\%) on precision (P), recall (R) and F1 score on the English test set of CoNLL-2009.}
    \label{tab:conll2009-results}

\vspace{0.3cm}

    \resizebox{1.0\linewidth}{!}{%
    \begin{tabular}{lcccc}
        \toprule
        
        
        
        \textbf{CoNLL-2012} & & \textbf{\textit{~P~}} & \textbf{\textit{~R~}} & \textbf{F1} \\
        
        \midrule
        
        \citet{li-etal-2019-dependency-or-span-srl} &&
        85.7 & 86.3 & 86.0 \\
        
        \citet{shi-lin-2019-simple-srl} && 85.9 & 87.0 & 86.5 \\
        
        \citeauthor{marcheggiani-titov-2020-graph} &&
        86.5 & 87.1 & 86.8 \\
        
        \citet{conia-navigli-2020-multilingual-srl} &&
        86.9 & 87.7 & 87.3 \\
        
        
        \citet{blloshmi-etal-2021-gsrl} &&
        87.8 & 86.8 & 87.3 \\
        
        This work$_\textrm{ CoNLL-2012}$ &&
        88.6 & 86.1 & 87.4 \\
        
        
        This work$_\textrm{ ALL}$ &&
        87.7 & 87.1 & 87.4 \\
        
        \bottomrule
        
    \end{tabular}
    }
    \caption{Results (\%) on precision (P), recall (R) and F1 score on the English test set of CoNLL-2012.}
    \label{tab:conll2012-results}

\vspace{0.3cm}

    \resizebox{1.0\linewidth}{!}{%
    \begin{tabular}{lcccc}
        \toprule
        
        
        
        \textbf{FrameNet} & & \textbf{\textit{~P~}} & \textbf{\textit{~R~}} & \textbf{F1} \\
        
        \midrule
        
        \citet{swayamdipta-etal-2017-open-sesame} &&
        70.5 & 66.7 & 68.6 \\
        
        \citet{peng-etal-2018-learning} &&
        80.2 & 72.9 & 76.4 \\
        
        \citeauthor{marcheggiani-titov-2020-graph} &&
        77.8 & 76.9 & 77.4 \\
        
        \citet{pancholy-etal-2021-sister} &&
        72.1 & 70.2 & 71.1 \\
        
        This work$_\textrm{ FrameNet}$ &&
        79.2 & 79.3 & 79.3 \\
        
        
        This work$_\textrm{ ALL}$ &&
        79.9 & 79.9 & 79.9 \\
        
        \bottomrule
        
    \end{tabular}
    }
    \caption{Results (\%) on precision (P), recall (R) and F1 score on the English test set of FrameNet.}
    \label{tab:framenet-results}
\end{table}

\paragraph{PropBank-style SRL.}
We first discuss the results obtained by DSRL on the gold standard benchmarks provided as part of the CoNLL-2009 and CoNLL-2012 Shared Tasks, annotated with PropBank sense and role labels.
As can be seen in Table~\ref{tab:conll2009-results}, we observe strong results in dependency-based SRL, reaching an F1 score of 92.5\% in the English test set of CoNLL-2009.
Therefore, despite having to cast our natural language descriptions to discrete labels, our approach performs in the same ballpark as the most recent state-of-the-art systems proposed by \citet{conia-navigli-2020-multilingual-srl} and \citet{blloshmi-etal-2021-gsrl}; the fact that our approach is able to slightly outperform the latter (+0.1\% in F1 score) is particularly meaningful, as they adopt the same pretrained language model (BART-large).
We can observe the same behavior in span-based SRL, where our model -- without any task-specific modifications -- marginally surpasses (+0.1\% in F1 score) that of \citet{blloshmi-etal-2021-gsrl} on the English test set of CoNLL-2012, as shown in Table~\ref{tab:conll2012-results}.
Thus, the key observation here is that a natural language output does not necessarily hurt performance.

\paragraph{FrameNet-style SRL.}
As shown in Appendix~\ref{app:definition-stats}, PropBank definitions for predicate senses and semantic roles are quite short, and therefore one may wonder whether our task reformulation is feasible in practice when using longer definitions from richer sources, such as FrameNet, in which the label definitions are up to three times longer.
From our experiments, this is, indeed, the case: our approach achieves state-of-the-art results in full-structure extraction \citep{baker-etal-2007-semeval} on the test set of FrameNet 1.7, obtaining 79.3 in F1 score (Table \ref{tab:framenet-results}).
We note that the results are not directly comparable with previous work, as DSRL employs a language model (BART) that is different from that of other approaches, e.g., \citet{marcheggiani-titov-2020-graph} used RoBERTa.
However, the results achieved by DSRL still indicate the performance that a generative approach can obtain in frame-semantic parsing~\cite{das-etal-2014-frame}, which might be considered more complex than PropBank-based SRL.
Indeed, predicates in FrameNet usually have a higher degree of polysemy, and the semantic roles are sparser, e.g., there are more than 2000 different semantic roles in FrameNet 1.7 compared to only 50-60 semantic roles in the PropBank releases used for the CoNLL-2009 and CoNLL-2012 shared tasks (see Table~\ref{tab:dataset-stats-train} in Appendix~\ref{app:data-stats}).

\begin{table}[t]
    \centering
    \resizebox{1.0\linewidth}{!}{%
    \begin{tabular}{lcccc}
        \toprule
        
        
        
         \textbf{CoNLL-2009 (OOD)} & & \textbf{\textit{~P~}} & \textbf{\textit{~R~}} & \textbf{F1} \\
        
        \midrule
        
        \citet{li-etal-2019-dependency-or-span-srl} &&
        -- & -- & 81.5 \\
        
        \citet{he-etal-2019-syntax-aware-srl} &&
        -- & -- & 82.2 \\
        
        \citet{shi-lin-2019-simple-srl} &&
        -- & -- & 92.4 \\
        
        \citet{conia-navigli-2020-multilingual-srl} &&
        -- & -- & 85.9 \\
        
        \citet{blloshmi-etal-2021-gsrl} &&
        85.8 & 84.5 & 85.2 \\
        
        This work$_\textrm{ CoNLL-2009}$ &&
        86.4 & 84.8 & 85.6 \\
        
        This work$_\textrm{ ALL}$ &&
        86.1 & 86.4 & 86.3 \\
        
        \bottomrule
        
    \end{tabular}
    }
    \caption{Results (\%) on precision (P), recall (R) and F1 score on the English out-of-domain test set of CoNLL-2009.}
    \label{tab:conll2009-ood-results}
\end{table}

\paragraph{Combining PropBank and FrameNet.}
The flexibility of our approach is evidenced by the fact that our model can benefit from learning to perform jointly dependency-based PropBank-style SRL on CoNLL-2009, span-based PropBank-style SRL on CoNLL-2012, and span-based FrameNet-style SRL on FrameNet 1.7, simply by enforcing two inventory-specific special tokens at the beginning of the decoding process, e.g., \texttt{\specials{propbank}\specials{dep-srl}} $t^p$ or \texttt{\specials{framenet}\specials{span-srl}} $t^p$, where $t^p$ is the target output, i.e., the semantically-augmented sentence described in Section~\ref{subsec:description-generation}.
Using natural language descriptions instead of discrete labels as the common denominator across heterogeneous inventories yields similar -- or even improved -- results when training our model on the three resources at the same time, compared to training a separate model on each dataset, as reported in the last row of Tables~\ref{tab:conll2009-results}, \ref{tab:conll2012-results}, and \ref{tab:framenet-results}, removing the need for separate systems for different setups and empirically supporting the flexibility of our model in scaling across dissimilar formalisms (dependency- and span-based annotations) and linguistic theories (PropBank and FrameNet).
Indeed, our model is able to leverage such features to achieve a new state of the art in the out-of-domain test set of CoNLL-2009.
As shown in Table~\ref{tab:conll2009-ood-results}, when we train DSRL jointly on CoNLL-2009, CoNLL-2012, and FrameNet, we can observe a large improvement, achieving 86.3\% in F1 score -- +0.7\% over training DSRL only on CoNLL-2009, and +1.1\% over \citet{blloshmi-etal-2021-gsrl} -- and setting a new state of the art on this out-of-domain benchmark, to the best of our knowledge.

\section{Quantitative Analysis}

\subsection{Rare and Unseen Senses}
\label{sec:rare-unseen}
The probability with which a word assumes one of its possible senses follows Zipf's distribution \cite{kilgarriff04}, and thus it is very skewed towards the most frequent senses.
Here, we analyze the bias that our system shows in predicting the most frequent predicate senses on the following partitions of the CoNLL-2009 and CoNLL-2012 test sets: 
\begin{itemize*}
    \item[\romannumeral 1)]  \textbf{MFS}, all the instances containing predicates that are annotated with their most frequent sense;
    \item[\romannumeral 2)] \textbf{LFS}, all the instances containing predicates that are not annotated with their most frequent sense;
    \item[\romannumeral 3)] \textbf{UNSEEN}, all the instances containing predicates that are annotated with a sense that is not present in the training set.
\end{itemize*}

\begin{table*}[t!]
    \centering
    
    \resizebox{1.0\linewidth}{!}{%
        \begin{tabular}{clccrccrccrccr}
            \toprule
            & \multicolumn{1}{c}{} &
            & \multicolumn{2}{c}{ALL} &
            & \multicolumn{2}{c}{MFS} &
            & \multicolumn{2}{c}{LFS} &
            & \multicolumn{2}{c}{UNSEEN}\\
            
            \cmidrule{4-5} \cmidrule{7-8} \cmidrule{10-11} \cmidrule{13-14}
            & \multicolumn{1}{c}{\small{Dataset}} &
            & \multicolumn{1}{c}{\small{F1}} & \multicolumn{1}{c}{\small{Support}} & &
             \multicolumn{1}{c}{\small{F1}} & \multicolumn{1}{c}{\small{Support}} & &
             \multicolumn{1}{c}{\small{F1}} & \multicolumn{1}{c}{\small{Support}} &  &
             \multicolumn{1}{c}{\small{F1}} & \multicolumn{1}{c}{\small{Support}}\\
            
            \midrule\parbox[t]{2mm}{\multirow{2}{*}{\rotatebox[origin=c]{90}{\textit{Pred.}}}} &
            CoNLL-2009 && 97.6 & \num{8986} && 98.9 & \num{8056} {\small (89.7\%)} && 87.0 & \num{838} {\small ~~(9.3\%)} && 80.4 & \num{92} {\small (1.0\%)} \\
            & CoNLL-2012 && 96.7 & \num{62002} && 98.5 & \num{50607} {\small (81.6\%)} && 89.2 & \num{11026} {\small (17.8\%)} && 75.3 & \num{369} {\small (0.6\%)} \\
            \midrule
            \parbox[t]{2mm}{\multirow{2}{*}{\rotatebox[origin=c]{90}{\textit{Arg.}}}} &
            CoNLL-2009 && 89.2 & \num{19946} && 89.8 & \num{17753} {\small (89.0\%)} && 87.2 & \num{1964} {\small ~~(9.8\%)} && 58.5 & \num{229} {\small (1.1\%)} \\
            & CoNLL-2012 && 87.3 & \num{145055} && 88.1 & \num{123769} {\small (85.3\%)} && 83.9 & \num{20350} {\small (14.0\%)} && 62.4 & \num{936} {\small (0.6\%)} \\
    
            \bottomrule
        \end{tabular}
    }
    \caption{Predicate and argument labeling scores on the test sets of CoNLL-2009 and CoNLL-2012. We report the performance (F1) on the most frequent senses (MFS), least frequent senses (LFS) and unseen senses (UNSEEN). Support indicates the number of instances (percentage) of the corresponding class.}
    \label{tab:mfs-lfs-unseen}
\end{table*}

As we can see from Table \ref{tab:mfs-lfs-unseen}, the performance of our system on predicate sense disambiguation is strong in the MFS partition -- more than 98.5\% in both CoNLL-2009 and CoNLL-2012 -- since the vast majority of predicates are annotated with their most frequent sense.
This bias justifies the difference in F1 score between the MFS and LFS partitions, i.e., --11.9\% and --9.3\% on CoNLL-2009 and CoNLL-2012, respectively.
As far as the UNSEEN partition is concerned, on the other hand, we observe that our approach seems to be capable of generating and retrieving senses that it has never seen at training time with a relatively low decrease in performance (--6.6\% and --13.9\% compared to the results on the LFS partition).
Interestingly, the results on argument labeling are comparable between MFS and LFS predicates.
However, there is still large room for improvement in the argument labeling of UNSEEN predicates, whose argument structure represents a more challenging zero-shot setting. 

\subsection{Data efficiency}
\label{sec:train-down-sampling}
\begin{figure}[t]
    \centering
    \includegraphics[width=1.0\columnwidth]{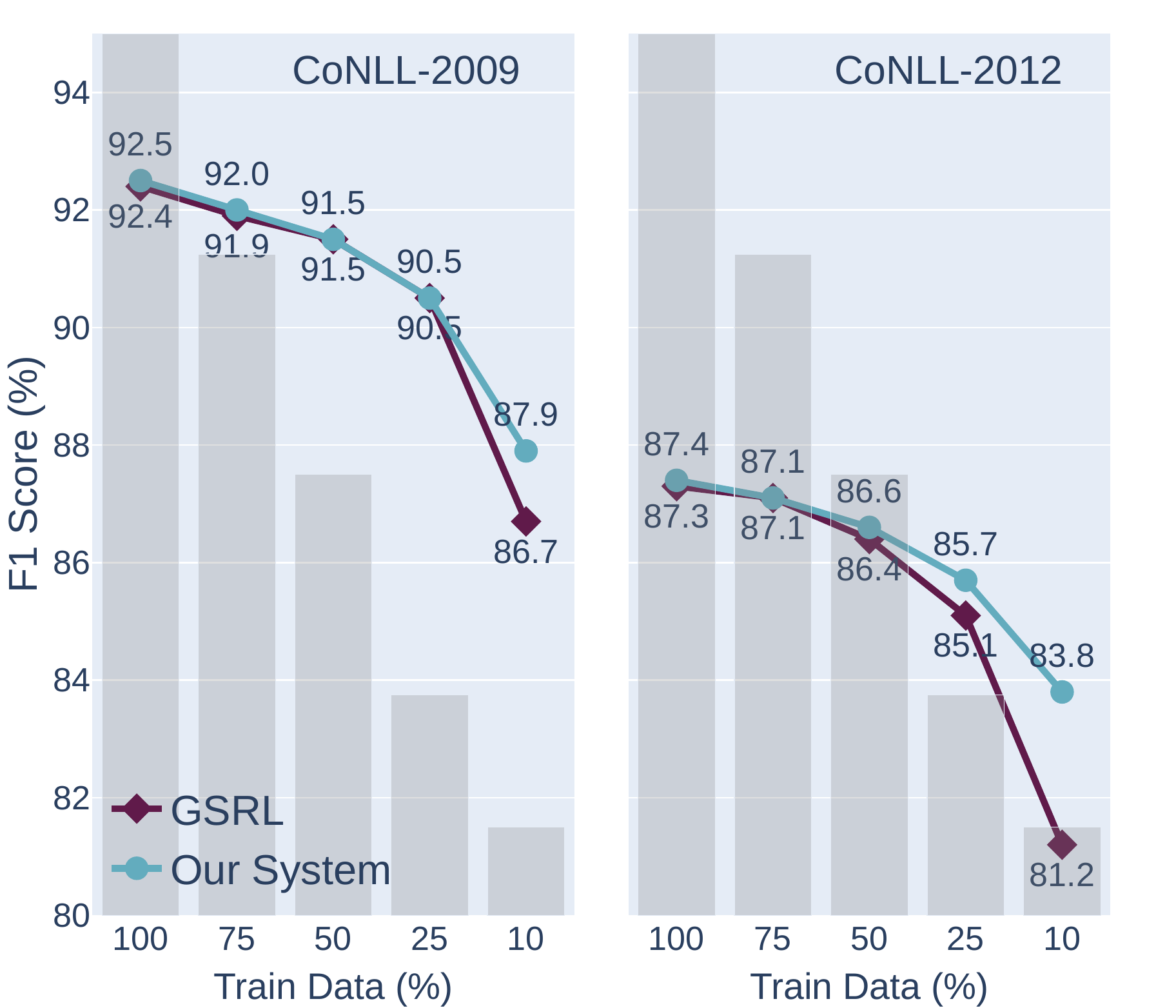}
    \caption{Performance comparison of our system and GSRL when down-sampling the training dataset to \num{10}\%, \num{25}\%, \num{50}\% and \num{75}\% of the total instances.}
    \label{fig:train-down-sampling}
\end{figure}

Considering the large expense entailed in manually annotating text with sense and role labels, we deem it indispensable to also evaluate the flexibility of a system in terms of its scalability on fewer training instances.
Therefore, we analyze the results of our model by gradually reducing the training set to \num{75}\%, \num{50}\%, \num{25}\%, and \num{10}\% of its original size, and compare this learning curve with that of GSRL~\citep{blloshmi-etal-2021-gsrl}.
Notwithstanding the significant differences between the two approaches, both show similar learning curves on CoNLL-2009 and CoNLL-2012 (Figure~\ref{fig:train-down-sampling}), confirming that manually annotating more sentences eventually ceases to provide large improvements: in fact, the enormous effort of doubling the training instances of CoNLL-2012 by annotating other 100,000 predicates (from 50\% to 100\% of its original size) results in less than a 1.0\% gain in F1 score.
Interestingly, our system shows higher data efficiency in the lowest data regime, especially for span-based SRL with a 2.6\% gain in F1 score over GSRL when they are both trained on 10\% of the original dataset.
We argue that our novel formulation better leverages the pretraining of the underlying language model in lower-data scenarios.
However, when more training data is available, task-specific approaches are eventually able to close the gap.

\begin{table}
\centering
\resizebox{1.0\linewidth}{!}{
\begin{tabular}{lccc}

\toprule

\textbf{Training Data (10\%)} & \textbf{\small CoNLL-09} & \textbf{\small CoNLL-12} & \textbf{\small FrameNet}\\

\midrule

CoNLL-2009 (C09) & 87.9 & -- & -- \\
CoNLL-2012 (C12) & -- & 83.8 & -- \\
FrameNet 1.7 (F17) & -- & -- & 74.9 \\

\midrule

C09 + C12 & 88.4 & 84.2 & -- \\
C12 + F17 & -- & 84.0 & 75.0 \\
C09 + F17 & 87.9 & -- & 75.1 \\

\midrule

C09 + C12 + F17 & 88.5 & 84.3 & 75.4 \\

\bottomrule

\end{tabular}
}
\caption{Results of our model when trained on a random sample of 10\% of the original training splits of CoNLL-2009, CoNLL-2012, and FrameNet 1.7 and their combinations.}
\label{tab:multi-dataset-few-shot}
\end{table}

Finally, we investigate whether our approach is still capable of handling multiple inventories at the same time in low-data regimes.
To this end, we trained the model with several combinations of inventories on 10\% of their training data.
As we can see from Table \ref{tab:multi-dataset-few-shot}, the model achieves improved results whenever it is trained on any two inventories, with the one trained jointly on CoNLL-2009, CoNLL-2012, and FrameNet performing best.
Interestingly enough, the model is able to handle the CoNLL-2009 + FrameNet combination despite the different linguistic resources (PropBank vs FrameNet) and annotation formalism (dependency- vs span-based SRL).

\section{Qualitative Analysis}
\label{sec:qualitative-analysis}

\subsection{Generation Examples}
In Table \ref{tab:generation-examples}, we provide some examples of the descriptions generated by our system.
Given an input sentence, we compare its gold standard sequence ($\hat{g}$) with the one generated automatically ($g$).
We find that, in some cases, the automatic descriptions are more contextual than the gold ones, occasionally overcoming the limitations of the linguistic inventories.
In Example 1, for instance, the gold definition of the predicate \textit{brandish.01} is only applicable to weapons; instead, the model-generated sequence is preferable as the entity brandished is a flag.
In other cases, such as in Example 2, our approach generates more descriptive definitions, e.g., \textit{depictor} instead of \textit{agent}, and \textit{thing described} rather than \textit{theme}.
Furthermore, we show some examples in which the model generates semantically-appropriate natural language descriptions for out-of-inventory, and thus unseen, predicates.  
Even in this setting, the model often generates semantically-appropriate natural language descriptions.
This is the case with Example 3, in which the model describes the semantics of \textit{nibble.01} (unseen at training time) by taking advantage of a similar predicate, namely, \textit{peck.01} (seen at training time).
This is also true for noun predicates, as shown in Example 4.

\begin{table}[t]
    \small
    \resizebox{\linewidth}{!}{%
    \begin{tabular}{lp{\linewidth}}
    \toprule
    
    \multicolumn{2}{l}{Ex. 1: Thousands of supporters, many \underline{brandishing} flags ...} \\
    \midrule
    \textbf{Gold} & brandish: wave weapons. ... brandishing [flags]\textcolor{Magenta}{\textbf{\{weapon\}}} ...\\[2pt]
    \textbf{Pred} & brandish: display, exhibit. ... brandishing [flags]\textcolor{Magenta}{\textbf{\{entity displayed\}}} ...\\
    
    \midrule
     
    \multicolumn{2}{l}{Ex. 2: [...] its unrealistic \underline{depiction} of the characters' [...] private lives.} \\
    \midrule
    \textbf{Gold} & depiction: show to be. ... [its]\textcolor{Magenta}{\textbf{\{agent\}}} [unrealistic]\{instrument or manner\} depiction [of]\textcolor{RoyalBlue}{\textbf{\{theme\}}} ...\\[2pt]
    \textbf{Pred} & depiction: show to be. ... [its]\textcolor{Magenta}{\textbf{\{depictor\}}} [unrealistic] \{instrument or manner\} depiction [of]\textcolor{RoyalBlue}{\textbf{\{thing described\}}} ... \\
  
    \midrule
    
    \multicolumn{2}{l}{Ex. 3: [...] he was "\underline{nibbling} at" selected stocks during Friday's plunge.} \\
    \midrule
    \textbf{Gold} & \textit{n/a} (out of inventory) \\[2pt]
    \textbf{Pred} & nibble: \textcolor{Turquoise}{\textbf{eat lightly}}. [he]\{eater\} was nibbling [at selected stocks]\{food\} [during Friday's plunge]\{time or duration\}. \\
    
    \midrule
    
    \multicolumn{2}{l}{Ex. 4: Zaire's president Mobutu met with [...] a senior U.S. \underline{envoy}.} \\
    \midrule
    \textbf{Gold} & \textit{n/a} (out of inventory) \\[2pt]
    \textbf{Pred} & envoy: \textcolor{Turquoise}{\textbf{stand for, correspond}}. ... a senior [U.S.]\{entity being substituted by the other\} [envoy]\{entity taking place of other\}. \\
    
    \bottomrule
    
    \end{tabular}
    }
    \caption{Generation examples. Given an input sentence, we compare the gold and the system-generated sequence. \underline{Predicates} are underlined.}
    \label{tab:generation-examples}
\end{table}

\subsection{Classes of Error}
We identify three main classes of error: 
the first is directly connected to our system (\textit{Disambiguation Errors}) and the other two (\textit{Out-of-Inventory Descriptions} and \textit{Retrieval Errors}) concern the noisy process we use to cast natural language descriptions to discrete class labels.

\paragraph{Disambiguation errors} occur when the model generates a definition that does not describe the correct sense of a predicate in a given context. 
For example, the system provides the wrong definition for the predicate ``bumble'' in the following sentence \textit{s}, misclassifying it as ``speak quietly'':
\begin{itemize}
    \vspace{-3pt}
    \item [$s$:] Shane survived the week only to have an executive \underline{bumbling} his way into a criminal investigation.
    \vspace{-6pt}
    \item \textcolor{ForestGreen}{\textbf{Gold:}} speak or move in a confused way
    \vspace{-10pt}
    \item \textcolor{BrickRed}{\textbf{Pred:}} speak quietly
\end{itemize}
We note that, given the autoregressive nature of the model, producing a wrong sense definition often compromises the entire argument structure.

\paragraph{Ouf-of-inventory descriptions} may be produced by our approach since it is not strictly tied to the vocabulary of a predefined linguistic resource.
While our model can generate predicate-argument structures not present in the inventory, they can still provide correct semantic explanations.
For instance, in the following sentence, the reference and the generated definitions convey the same semantics:
\begin{itemize}
    \item \textcolor{ForestGreen}{\textbf{Gold:}} dupe: \textit{trick}. He meets [a French girl]\{\textit{tricker}\} who \underline{dupes} [him]\{\textit{tricked}\} [into providing a home for her pet and then steals his car]\{\textit{induced action}\}.
    \vspace{-6pt}
    \item \textcolor{BrickRed}{\textbf{Pred:}} dupe: \textit{deceive}. He meets [a French girl]\{\textit{deceiver}\} who \underline{dupes} [him]\{\textit{victim}\} [into providing a home for her pet and then steals his car]\{\textit{tricked into}\}.
\end{itemize}
Associating ``victim'' to ``tricked'' is far from trivial, and such cases often result in \textbf{retrieval errors}, i.e., errors that are caused by the inability of the sentence embedding model -- SimCSE in our case -- to correctly capture the semantic similarity between the gold and generated definitions.

\section{Conclusion}
Recent progress in SRL has mainly revolved around the development of state-of-the-art systems which, however, are bound to specific predicate-argument inventories.
In this paper, instead, we proposed a novel task formulation that takes a step towards putting interpretability and flexibility in the foreground: we reframed SRL as the task of describing the predicate-argument structure of a sentence using natural language only, which is human-interpretable by definition.
Our experiments, supported by in-depth analyses, demonstrated that prioritizing interpretability does not come at the expense of performance.
Furthermore, our approach is flexible enough to achieve competitive or even state-of-the-art results on popular gold standard benchmarks for SRL, showing that natural language can act as a bridge between heterogeneous linguistic resources, e.g., PropBank and FrameNet, and also annotation formalisms, e.g., dependency- or span-based SRL.
We hope that our model will foster research in high-performance yet interpretable systems in NLP, and provide a means towards achieving easier integration of sentence-level semantics into downstream applications.

\section*{Limitations}

\paragraph{Generation.}
Although our model achieves results on gold standard benchmarks that are on par or even better than the current state of the art, its generative nature certainly makes it slower than previous work based on discriminative approaches \cite{he-etal-2019-syntax-aware-srl,shi-lin-2019-simple-srl,conia-etal-2021-unifying-srl}.
Indeed, our model generates the entire semantically-augmented sentence, i.e., the input sentence with its predicate-argument structures in natural language, autoregressively.
While this issue also affects our most direct competitor \cite{blloshmi-etal-2021-gsrl}, which generates discrete labels, this is still a limitation -- or, more precisely, a weakness -- we would like to remark.
Indeed, before deploying our system in production environments, one should carefully weigh the advantages of our method against its slower inference times.
The degree of slowdown will inevitably depend on the hardware, but we estimate that a generative approach could be several times slower than a discriminative one.
However, this could also be a matter for further research on the topic; for example, non-autoregressive generative models are steadily narrowing the performance gap \cite{gu-tan-2022-non} while mitigating the weaknesses of current autoregressive approaches.

\paragraph{Evaluation.}
Section~\ref{sec:qualitative-analysis} and Table~\ref{tab:generation-examples} provide a qualitative analysis of the behavior of our proposed approach on out-of-inventory instances, which may also include rare predicates or neologisms.
We acknowledge that a quantitative analysis of how our model really performs on out-of-inventory instances would provide sounder evidence of the benefits of our approach.
However, we do not possess the economic and human resources required to create a benchmark large enough for this purpose.
We believe that such a benchmark could be a great contribution to the area of SRL, but the endeavor of annotating a significant number of out-of-inventory instances will require further study.

\paragraph{Multilinguality.}
Extending our work to multiple languages is still a challenge and may require more effort than current approaches, such as that proposed by \citet{conia-etal-2021-unifying-srl} which uses language-specific decoders on top of a shared cross-lingual encoder.
One could consider pursuing a similar strategy, i.e., using a shared cross-lingual encoder and multiple language-specific autoregressive decoders.
However, the main limitation here is the availability and the structure of current linguistic inventories in other languages and, therefore, definitions in languages other than English.
For instance, the Chinese PropBank inventory provided as part of the CoNLL-2009 Shared Task lacks definitions for the majority of the predicate senses, whereas the latest version is not freely distributed.
Fortunately, the attention to multilingual SRL is increasing; for example, it would certainly be interesting to analyze the feasibility of our approach to the recently released global FrameNet project.




\section*{Ethics Statement}
Pretrained language models have been shown to manifest undesirable biases, inherited from the corpora on which they have been trained using self-supervision strategies.
We train our model starting from the weights of BART~\cite{lewis-etal-2020-bart} and, therefore, there is a high probability that these biases are also inherited, or even exaggerated, by our final models.
However, we did not investigate such biases in this work; hence, we advise against using our model in a production environment without a careful analysis beforehand.
Finally, we remark that the test sets of CoNLL-2009, CoNLL-2012, and FrameNet 1.7 also contain relatively old documents about economics, politics, and past events that do not reflect the current situation.
Therefore, the results of such benchmarks are intended only as a basis for comparison with previous approaches and not as a measure of the performance of our model in real-world applications.

\section*{Acknowledgments}
\begin{center}
\noindent
    \begin{minipage}{0.1\linewidth}
        \begin{center}
            \includegraphics[scale=0.2]{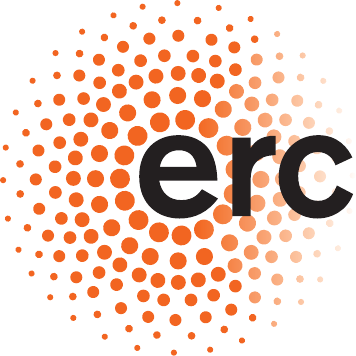}
        \end{center}
    \end{minipage}
    \hspace{0.01\linewidth}
    \begin{minipage}{0.70\linewidth}
        The authors gratefully acknowledge the support of the ERC Consolidator Grant MOUSSE No.\ 726487 under the European Union's Horizon 2020 research and innovation programme.
    \end{minipage}
    \hspace{0.01\linewidth}
    \begin{minipage}{0.1\linewidth}
        \begin{center}
            \includegraphics[scale=0.08]{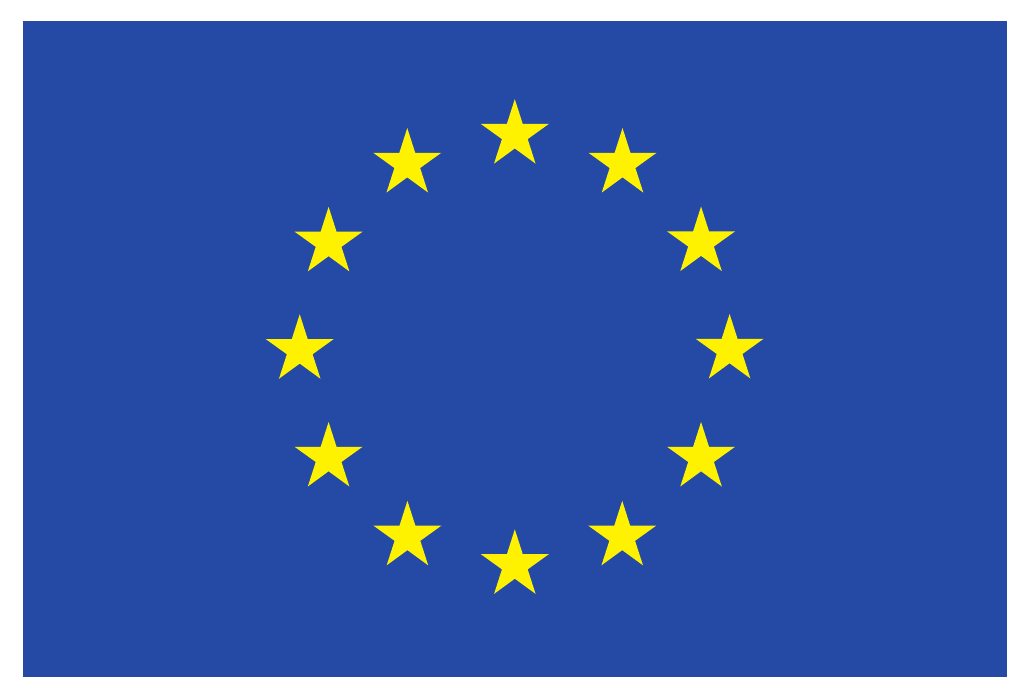}
        \end{center}
    \end{minipage}\\
\end{center}
\vspace{0.2cm}
\noindent This work was supported in part by the MIUR under grant ``Dipartimenti di Eccellenza 2018-2022'' of the Department of Computer Science of Sapienza University of Rome.


\bibliography{main}
\bibliographystyle{acl_natbib}

\appendix
\newpage
\section{Data License}
\label{app:data-license}
Both the CoNLL-2009 and CoNLL-2012 datasets are distributed by the Linguistic Data Consortium (LDC) and can be used under the LDC license.\footnote{\url{https://www.ldc.upenn.edu/data-management/using/licensing}}
FrameNet 1.7 -- the linguistic resource and its annotated dataset -- is freely available upon request.\footnote{\url{https://framenet.icsi.berkeley.edu/fndrupal}}
We note that the original Shared Task of CoNLL-2012 was concerned with the task of Coreference Resolution; however, given its SRL annotations, it soon also became a popular benchmark for span-based SRL.

\section{Data Statistics}
\label{app:data-stats}
In Tables \ref{tab:dataset-stats-train}, \ref{tab:dataset-stats-validation}, and \ref{tab:dataset-stats-test}, we provide an overview of the statistics of the train, validation and test sets, respectively, for the datasets we use in our experiments, namely,  the English splits of CoNLL-2009, CoNLL-2012, and FrameNet 1.7.
In particular, for each dataset, we report the number of sentences and their average length in tokens, with FrameNet having the longest sentences on average (+20\% over CoNLL-2009 and +40\% over CoNLL-2012).
We also report the number of annotated predicates for each dataset; interestingly, each predicate in FrameNet features around 6 arguments per predicate, a value that is much larger than those of CoNLL-2009 and CoNLL-2012, which feature around 2.5 arguments per predicate.
These are probably the reasons why the FrameNet dataset is particularly challenging, even for modern neural-based models.

Finally, we can also appreciate the heterogeneity between the characteristics of PropBank-style and FrameNet-style SRL.
Indeed, FrameNet clusters predicate senses into frames, resulting in a smaller number of predicate classes (around 1,000) compared to PropBank (5,000 to 8,000).
At the same time, the frame-specific semantic roles of FrameNet result in a much larger number of role classes compared to the coarse-grained semantic roles of PropBank.

\begin{table*}[t!]
    \centering
    \begin{tabular*}{\textwidth}{l@{\extracolsep{\fill}}rrrrrrrr}
        \toprule
        
        & \multicolumn{4}{c}{Sentences}
        & \multicolumn{2}{c}{Predicates}
        & \multicolumn{2}{c}{Arguments}\\
        
        \cmidrule{2-5} \cmidrule{6-7} \cmidrule{8-9}
        
        & \multicolumn{1}{c}{\small{Total$_s$}} & \multicolumn{1}{c}{\small{Distinct$_s$}} & \multicolumn{1}{c}{\small{Annotated}} & \multicolumn{1}{c}{\small{Avg. Len.}}
        & \multicolumn{1}{c}{\small{Total$_p$}} & \multicolumn{1}{c}{\small{Senses}} 
        & \multicolumn{1}{c}{\small{Total$_a$}} & \multicolumn{1}{c}{\small{Roles}}\\
        
        \midrule
        CoNLL-2009 & 39,279 & 38,770 & 37,847 & 24.4 & 179,014 & 8,237 & 393,699 & 52\\
        CoNLL-2012 & 115,812 & 109,374 & 90,856 & 19.0 & 253,070 & 5,287 & 598,983 & 66\\
        FrameNet & 19,391 & 3,353 & 19,391 & 29.5 & 20,105 & 859 & 123,977 & 2,042 \\

        \bottomrule
    \end{tabular*}
    \caption{Overview of the CoNLL-2009, CoNLL-2012, and FrameNet training datasets. For each dataset we report the number of sentences (\textit{Total$_s$}), the number of sentences with at least an annotated predicate (\textit{Annotated}), the average number of tokens per sentence (\textit{Avg. Len.}), the number of predicates (\textit{Total$_p$}) and predicate senses (\textit{Senses}), and also the number of arguments (\textit{Total$_a$}) and argument roles (\textit{Roles}).}
    \label{tab:dataset-stats-train}

\vspace{5mm}

    \centering
    \begin{tabular*}{\textwidth}{l@{\extracolsep{\fill}}rrrrrrrr}
        \toprule
        
        & \multicolumn{4}{c}{Sentences}
        & \multicolumn{2}{c}{Predicates}
        & \multicolumn{2}{c}{Arguments}\\
        
        \cmidrule{2-5} \cmidrule{6-7} \cmidrule{8-9}
        
        & \multicolumn{1}{c}{\small{Total$_s$}} & \multicolumn{1}{c}{\small{Distinct$_s$}} & \multicolumn{1}{c}{\small{Annotated}} & \multicolumn{1}{c}{\small{Avg. Len.}}
        & \multicolumn{1}{c}{\small{Total$_p$}} & \multicolumn{1}{c}{\small{Senses}} 
        & \multicolumn{1}{c}{\small{Total$_a$}} & \multicolumn{1}{c}{\small{Roles}}\\
        
        \midrule
        CoNLL-2009 & 1,334 & 1,334 & 1,283 & 25.0 & 6,390 & 1,990 & 13,865 & 32\\
        CoNLL-2012 & 15,680 & 15,086 & 12,600 & 19.4 & 35,297 & 2,912 & 83,362 & 48\\
        FrameNet & 2,272 & 326 & 2,272 & 35.2 & 2,382 & 394 & 17,347 & 893 \\

        \bottomrule
    \end{tabular*}
    \caption{Overview of the CoNLL-2009, CoNLL-2012, and FrameNet validation datasets. For each dataset we report the number of sentences (\textit{Total$_s$}), the number of sentences with at least an annotated predicate (\textit{Annotated}), the average number of tokens per sentence (\textit{Avg. Len.}), the number of predicates (\textit{Total$_p$}) and predicate senses (\textit{Senses}), and also the number of arguments (\textit{Total$_a$}) and argument roles (\textit{Roles}).}
    \label{tab:dataset-stats-validation}

\vspace{5mm}

    \centering
    \begin{tabular*}{\textwidth}{l@{\extracolsep{\fill}}rrrrrrrr}
        \toprule
        
        & \multicolumn{4}{c}{Sentences}
        & \multicolumn{2}{c}{Predicates}
        & \multicolumn{2}{c}{Arguments}\\
        
        \cmidrule{2-5} \cmidrule{6-7} \cmidrule{8-9}
        
        & \multicolumn{1}{c}{\small{Total$_s$}} & \multicolumn{1}{c}{\small{Distinct$_s$}} & \multicolumn{1}{c}{\small{Annotated}} & \multicolumn{1}{c}{\small{Avg. Len.}}
        & \multicolumn{1}{c}{\small{Total$_p$}} & \multicolumn{1}{c}{\small{Senses}} 
        & \multicolumn{1}{c}{\small{Total$_a$}} & \multicolumn{1}{c}{\small{Roles}}\\
        
        \midrule
        CoNLL-2009 & 2,000 & 1,999 & 1,913 & 24.4 & 8,987 & 2,254 & 19,949 & 35\\
        CoNLL-2012 & 27,897 & 26,698 & 21,863 & 19.2 & 62,012 & 3,489 & 145,078 & 50\\
        FrameNet & 6,714 & 1,247 & 6,714 & 27.2 & 6,872 & 620 & 34,454 & 1,354 \\

        \bottomrule
    \end{tabular*}
    \caption{Overview of the CoNLL-2009, CoNLL-2012, and FrameNet test datasets. For each dataset we report the number of sentences (\textit{Total$_s$}), the number of sentences with at least an annotated predicate (\textit{Annotated}), the average number of tokens per sentence (\textit{Avg. Len.}), the number of predicates (\textit{Total$_p$}) and predicate senses (\textit{Senses}), and also the number of arguments (\textit{Total$_a$}) and argument roles (\textit{Roles}).}
    \label{tab:dataset-stats-test}
\end{table*}

\section{Training Sequence Statistics}
\label{sec:appendix_output_sequences}
In Table \ref{tab:sequences_length}, we report the average length in characters of the sequences used to train our model.
As we can see, FrameNet 1.7 features the longest sequences among the three datasets we take into account, in line with what we report in Appendix~\ref{app:data-stats}.

\begin{table}[t]
\centering
\begin{tabular*}{\columnwidth}{@{\extracolsep{\fill}}lc@{\extracolsep{\fill}}}

\toprule
\multirow{2}{*}{\textbf{Dataset}} & \textbf{Avg. Len.}\\
& \textbf{\small (characters)}\\

\midrule

CoNLL-2009 & ~~83.1 \\
CoNLL-2012 & 127.6 \\
FrameNet 1.7 & 205.3 \\

\bottomrule
\end{tabular*}
\caption{CoNLL-2009, CoNLL-2012, and FrameNet training sequence statistics. For each dataset, we report the average length in characters of the sequence used for training the model.}
\label{tab:sequences_length}
\end{table}

\section{Argument Modifiers Definitions}
\label{app:argument-modifiers}
The English PropBank features two categories of semantic roles: core and adjunct.
If we define a semantic role as the relationship between an action or event (predicate) and one of the participants (argument), then the former category includes all those semantic roles that mark an important participant in the event, one that is expected to take part in it.
In PropBank, these core roles are identified using the labels \textsc{Arg0}, \textsc{Arg1}, $\dots$, \textsc{Arg5}, and their definitions change from predicate sense to predicate sense.
Instead, the second category, namely the adjunct roles or argument modifiers, are general roles whose semantics is not specific to a particular predicate and, therefore, can be used to tag general arguments, e.g., the time of the action (\textsc{ArgM-Tmp}) or the place of the event (\textsc{ArgM-Loc}).
We use the PropBank guidelines to translate such labels into natural language.
In Tables~\ref{tab:conll2009_argm_definitions} and \ref{tab:conll2012_argm_definitions}, we list the argument modifiers definitions that we use to train our model on CoNLL-2009 and CoNLL-2012, respectively.

While we aimed at creating argument modifier definitions that are homogeneous with the core role definitions, we remark that we did not perform a search for better definitions.
As one can see, some of the definitions reported in Tables~\ref{tab:conll2009_argm_definitions} and \ref{tab:conll2012_argm_definitions} are the natural language equivalent of the labels (e.g, \textsc{ArgM-Adv} and its definition ``adverbial modifier'', \textsc{ArgM-Lvb} and its definition ``light verb'', or \textsc{ArgM-Prd} and its definition ``secondary predication'', among others).
We believe that a possible venue for future research is looking into how we can create better definitions for such semantic roles.

\begin{table}[t]
\centering
\begin{tabular}{ll}
\toprule
\textbf{Argument Modifier} & \textbf{Definition}\\
\midrule
     AM-ADV &  adverbial modifier\\
     AM-CAU &  cause or reason \\
     AM-DIR &  direction or source \\
     AM-DIS &  discourse connective \\
     AM-EXT &  amount or extent \\
     AM-LOC &  location or position \\
     AM-MNR &  instrument or manner \\
     AM-MOD &  modal auxiliary \\
     AM-NEG &  negation marker \\
     AM-PNC &  purpose, not cause \\
     AM-PRD &  secondary predication \\
     AM-TMP &  time or duration \\
    \bottomrule
     \end{tabular}
\caption{CoNLL-2009 argument modifiers definitions. We provide descriptions for argument modifiers when they are not specified in the given predicate roleset.}
\label{tab:conll2009_argm_definitions}
\end{table}


\begin{table}[t]
\centering
\begin{tabular}{ll}
\toprule
\textbf{Argument Modifier} & \textbf{Definition}\\
\midrule
     ARGM-ADJ &  adjectival modifier  \\
     ARGM-ADV &  adverbial modifier  \\
     ARGM-CAU &  cause or reason  \\
     ARGM-COM &  comitative  \\
     ARGM-DIR &  direction or source  \\
     ARGM-DIS &  discourse connective  \\
     ARGM-EXT &  amount or extent  \\
     ARGM-GOL &  goal or destination  \\
     ARGM-LOC &  location or position  \\
     ARGM-LVB &  light verb  \\
     ARGM-MNR &  instrument or manner  \\
     ARGM-MOD &  modal auxiliary  \\
     ARGM-NEG &  negation marker  \\
     ARGM-PNC &  purpose, not cause  \\
     ARGM-PRD &  secondary predication  \\
     ARGM-PRP &  purpose or motivation  \\
     ARGM-TMP &  time or duration \\
     \bottomrule
\end{tabular}
\caption{CoNLL-2012 argument modifiers definitions. We provide descriptions for argument modifiers when they are not specified in the given predicate roleset.}
\label{tab:conll2012_argm_definitions}
\end{table}

\begin{table*}[t]
    \centering
    \begin{tabular*}{\textwidth}{l@{\extracolsep{\fill}}rrrrrr}
        \toprule
        
        & \multicolumn{2}{c}{Train}
        & \multicolumn{2}{c}{Validation}
        & \multicolumn{2}{c}{Test}\\
        
        \cmidrule{2-3} \cmidrule{4-5} \cmidrule{6-7}
        
        & \multicolumn{1}{c}{\small{Distinct$_d$}} & \multicolumn{1}{c}{\small{Avg. Len.$_d$}} &
         \multicolumn{1}{c}{\small{Distinct$_d$}} & \multicolumn{1}{c}{\small{Avg. Len.$_d$}} &
         \multicolumn{1}{c}{\small{Distinct$_d$}} & \multicolumn{1}{c}{\small{Avg. Len.$_d$}}\\
        
        \midrule
        CoNLL-2009 & 1,317 & 16.0 & 1,207 & 16.5 & 1,317 & 16.0 \\
        CoNLL-2012 & 4,401 & 19.5 & 2,393 & 18.4 & 2,864 & 18.7\\
        FrameNet & 3,750 & 46.7 & 882 & 47.3 & 1,982 & 48.1 \\

        \bottomrule
    \end{tabular*}
    \caption{CoNLL-2009, CoNLL-2012, and FrameNet predicate definitions statistics. For each dataset and split we report the number of distinct definitions (\textit{Distinct$_d$}), and their average length in characters (\textit{Avg. Len.$_d$}).}
    \label{tab:predicate-definitions-stats}

\vspace{5mm}

    \centering
    \begin{tabular*}{\textwidth}{l@{\extracolsep{\fill}}rrrrrr}
        \toprule
        
        & \multicolumn{2}{c}{Train}
        & \multicolumn{2}{c}{Validation}
        & \multicolumn{2}{c}{Test}\\
        
        \cmidrule{2-3} \cmidrule{4-5} \cmidrule{6-7}
        & \multicolumn{1}{c}{\small{Distinct$_d$}} & \multicolumn{1}{c}{\small{Avg. Len.$_d$}} &
         \multicolumn{1}{c}{\small{Distinct$_d$}} & \multicolumn{1}{c}{\small{Avg. Len.$_d$}} &
         \multicolumn{1}{c}{\small{Distinct$_d$}} & \multicolumn{1}{c}{\small{Avg. Len.$_d$}}\\
        
        \midrule
        CoNLL-2009 & 1,255 & 15.6 & 1,161 & 15.4 & 1,255 & 15.6 \\
        CoNLL-2012 & 5,002 & 16.9 & 2,477 & 16.4 & 3,032 & 16.4\\
        FrameNet & 2,167 & 58.7 & 634 & 58.2 & 1,184 & 57.0 \\

        \bottomrule
    \end{tabular*}
    \caption{CoNLL-2009, CoNLL-2012, and FrameNet role definitions statistics. For each dataset and split we report the number of distinct definitions (\textit{Distinct$_d$}), and their average length in characters (\textit{Avg. Len.$_d$}).}
    \label{tab:role-definitions-stats}
\end{table*}

\section{Definitions Statistics}
\label{app:definition-stats}
The length of the sequence that our model generates in output is certainly dependent on the length of the definitions we use to describe the sense of a predicate and its arguments.
In this Appendix, we provide a broad look at the number of unique sense and role definitions that appear in the train, validation, and test sets of CoNLL-2009, CoNLL-2012 and FrameNet 1.7.

As we can see in Table~\ref{tab:predicate-definitions-stats}, even though CoNLL-2009 and CoNLL-2012 are both tagged using PropBank labels, the number of distinct predicate sense definitions is quite different between the two datasets (1,317 unique definitions in the training set of CoNLL-2009 against 4,401 in CoNLL-2012).
This difference is probably due to the narrower domain of CoNLL-2009, which features a significant portion of sentences about finance from the Wall Street Journal, whereas CoNLL-2012 covers a more varied set of domains.
Although the number of unique sense definitions is different, the average length of these definitions between CoNLL-2009 and CoNLL-2012 is close, suggesting homogeneous definitions despite the use of two different versions of the English PropBank.
This is not the case when comparing the average length of the PropBank definitions used for CoNLL-2009 and CoNLL-2012 with those of FrameNet.
Indeed, predicate sense definitions in FrameNet are two to three times longer on average than PropBank's.
However, the experimental results reported in Tables~\ref{tab:framenet-results} and \ref{tab:multi-dataset-few-shot} show that our proposed generative model is still able to produce longer sense definitions.

We can observe a similar picture in Table~\ref{tab:role-definitions-stats} for the definitions of the semantic roles.
Interestingly, the difference between CoNLL-2009 and CoNLL-2012 in the average length of the semantic role definitions is even narrower, whereas the difference in length between PropBank-style and FrameNet-style role definitions widens even further, with FrameNet using role definitions that are almost four times longer than PropBank's.
The difference in length between the predicate sense and semantic role definitions between FrameNet and PropBank can be explained by the fact that, in the former resource, the definitions are richer and more detailed.
For example, the agent of the predicate \textit{provide} is defined just as ``giver'' in PropBank, whereas in FrameNet is defined as ``person that begins in possession of the theme and causes it to be in the possession of the recipient''.

\begin{table*}[t]
\centering
\begin{tabular}{lll}
\toprule
\textbf{Used in} & \textbf{Special Token(s)} & \textbf{Description}\\
\midrule
    Input & \texttt{<p></p>} &  indicate the start/end of a predicate \\
    Output &  \texttt{<reference-to>} & argument referring to another one (e.g., R-Arg1) \\
    Output & \texttt{<continuation-of>} & continuation of another argument (e.g., C-Arg1) \\
    Output &  \texttt{<propbank>} & Instructs the model to perform PropBank-style SRL \\
    Output &  \texttt{<framenet>} & Instructs the model to perform FrameNet-style SRL \\
    Output &  \texttt{<span-srl>} & Instructs the model to perform span-based SRL \\
    Output &  \texttt{<dep-srl>} & Instructs the model to perform dependency-based SRL \\
\bottomrule
\end{tabular}
\caption{List of the special tokens and their use in our model. For each special token, we indicate whether it is used in the input or in the output sequence. Some of these special tokens can be used in combination, e.g., \texttt{<propbank>}\texttt{<dep-srl>} to instruct the model to perform PropBank-style dependency-based SRL.}
\label{tab:special_tokens}

\vspace{5mm}

\centering
\resizebox{1.0\linewidth}{!}{%

\begin{tabular}{ll}
\toprule
\textbf{Special Token} & \textbf{Example}\\
\midrule
     \texttt{<p></p>} &  Not all those who \texttt{\textbf{<p>}} wrote \texttt{\textbf{</p>}} oppose the changes. \\
     
     \texttt{<p></p>} &  A \texttt{\textbf{<p>}} record \texttt{\textbf{</p>}} date has not been set. \\
     
     \texttt{<reference-to>} & ...It was [during this year]\{time or duration\} [that]\{ \texttt{\textbf{<reference-to>}} time or duration\} [the Japanese... \\
     
     \texttt{<continuation-of>} & [Japan]\{helper\}, [in terms of ...]\{ adverbial modifier\} , [it]\{\texttt{\textbf{<continuation-of>}} helper\} could have helped...\\
     
     \bottomrule
     \end{tabular}
     }
\caption{Examples of how the special tokens are inserted in the input or output sequence.}
\label{tab:special_tokens_examples}
\end{table*}

\section{Special Tokens}
\label{app:special-tokens}
As mentioned in Section~\ref{subsec:description-generation}, we use some special tokens to instruct the model on some task-specific functions.
For example, we pre-identify a predicate in an input sentence by surrounding its tokens with the special tokens \texttt{<p>} and \texttt{</p>}, indicating the start and the end of a predicate, respectively.
Table~\ref{tab:special_tokens} lists all the special tokens we use in our model in addition to the standard ones (e.g., \texttt{<s>} and \texttt{</s>} to indicate the start and end of the generated sequence).

We note that some of these special tokens can be used in combination.
For example, combining \texttt{<propbank}\texttt{<span-srl>} informs the model that we want it to generate a sentence annotated with PropBank-style definitions according to the span-based formalism; instead, combining  \texttt{<framenet}\texttt{<span-srl>} will result in a sentence annotated with FrameNet-style definitions using a span-based formalism.

For reference, we also provide a few examples of how these special tokens are inserted in an input or output sequence in Table~\ref{tab:special_tokens_examples}, using sentences from the training set of CoNLL-2012.

For the implementation, we simply add these special tokens to the input and output vocabulary of the underlying language model (i.e., BART).
The embeddings corresponding to the special tokens are randomly initialized and updated during training.

\end{document}